# Image Splicing Detection Using Inherent Lens Radial Distortion

H. R. Chennamma[1], Lalitha Rangarajan[2]

[1] Department of Studies in Computer Science, University of Mysore
Mysore-570 006, INDIA
*anusha_hr@rediffmail.com*

[2] Department of Studies in Computer Science, University of Mysore
Mysore-570 006, INDIA
*lali85arun@yahoo.co.in*

**Abstract**
Image splicing is a common form of image forgery. Such alterations may leave no visual clues of tampering. In recent works camera characteristics consistency across the image has been used to establish the authenticity and integrity of digital images. Such constant camera characteristic properties are inherent from camera manufacturing processes and are unique. The majority of digital cameras are equipped with spherical lens and this introduces radial distortions on images. This aberration is often disturbed and fails to be consistent across the image, when an image is spliced. This paper describes the detection of splicing operation on images by estimating radial distortion from different portions of the image using line-based calibration. For the first time, the detection of image splicing through the verification of consistency of lens radial distortion has been explored in this paper. The conducted experiments demonstrate the efficacy of our proposed approach for the detection of image splicing on both synthetic and real images.

## 1. Introduction

As digital technology advances, the need for authenticating digital images, validating their content and detection of forgeries has also increased. Common manipulations on images are copying and pasting portions of the image onto the same or another image to create a composite image. Proving the authenticity and integrity of an image is a challenging task. There are two common properties that an untampered image must have: natural scene quality and natural imaging quality [1]. For example, an image with inconsistency between the light direction and the shadow is not authentic because it fails to satisfy natural scene quality. Any output image naturally inherits the characteristic properties of the acquisition device. An image does not meet the natural imaging quality if different parts of the image do not share consistent characteristics of imaging device. A skilled forger can manage to satisfy the natural scene quality by using sophisticated image editing software but natural imaging quality is very difficult to achieve. This motivates us to take up this research.

The aim of this research is to demonstrate that it is possible to use inherent lens aberrations as unique fingerprints in the images for the detection of image splicing. Inconsistency in the degree of lens distortion in different portions of the image leads to the detection of image tampering. It is generally accepted that the optics of most consumer level cameras deviate from the ideal pinhole camera model. Among different kinds of aberrations, lens radial distortion is the most severe. The lens radial distortion causes non-linear geometrical distortion on images. In this paper we propose a novel passive technique (with no watermark or signature) for detecting copy-paste forgery by quantitatively measuring lens radial distortion across the image using line-based calibration. We estimate lens radial distortion from straight edges projected on images. Hence neither calibration pattern nor information about other camera parameter is necessary. Usually a composite image is created not just by weaving different portions of the same or different images, but it is also accompanied by subsequent operations like JPEG compression, contrast/brightness adjustment, color changing, blurring, rotation, resizing etc. to hide the obvious traces of tampering.

The remainder of this paper is organized as follows. Section 2 reviews the relevant research on image splicing detection. Section 3 details the background of lens radial distortion and also describes how to estimate lens radial distortion parameter for each detected line segment in the image. Section 4 analyses the effect of lens distortion at various zoom levels, for different cameras and also presents experimental results on the detection of image tampering performed on both synthetic images and real images. Section 5 discusses the limitations of the proposed method and future work. Section 6 concludes the paper.

## 2. Related Work

Digital image forensics is emerging as an interesting and challenging field of research [2, 3] recently. Here we present the review of related works on image splicing detection.

One of the kinds of image tampering is object removal where the regions of unwanted objects in an image are replaced by other parts of the same image. This type of operation is called copy-move or region-duplication. Methods in [4-6] are specifically designed to detect region duplication and are all based on block matching. First, the method divides an image into small blocks. Then it extracts the features from each block and hence, identifies possible duplicated regions on comparison. The main difference of these methods is the choice of features. Fridrich, et al. [4] have analyzed the DCT coefficients from each block. Popescu, et al. [5] have employed the principal component analysis (PCA) to reduce the image blocks into a PCA feature vector. Luo et al. [6] have extracted seven features in each block. Their experimental results demonstrated that the method could resist more post-processing operations.

Another kind of image tampering is splicing. Unlike region duplication, image splicing is defined as a simple joining of fragments of two or more different images. Several researchers have investigated the problem of splicing based on statistical properties of pixels (called pixel-based techniques) and camera characteristics (called camera-based techniques). Now, let us briefly review the literature on both techniques.

Johnson et al. [7] have described a method for revealing, traces of tampering using light inconsistencies as it is often difficult to match the lighting conditions from the individual photographs. Tian-Tsong Ng et al. [8] have described techniques to detect photomontaging. They have designed a classifier based on bi-coherence features of the natural images and photomontaged images. They have also proposed a mathematical model for image splicing [9]. One of the fundamental operations that need to be carried out in order to create forgeries is resizing (resampling). It is an operation that is likely to be carried out irrespective of the kind of forgery (copy move, photomontage, etc). Farid et al. [11] have described a method for estimation of resampling parameters in a discrete sequence and have shown its applications to image forensics. Chen et al. [12] have analyzed phase congruency for detection of image splicing. We have explored the use of wavelets for the detection of resampled portions [13].

Methods for the detection of image alteration based on inherent characteristics of digital camera (camera-based techniques) have been reported in [14-18]. Johnson et al. [14] have explored lateral chromatic aberration as a tool for detecting image tampering. Lateral chromatic aberration manifests itself, to a first-order approximation, as an expansion/contraction of color channels with respect to one another. When tampering with an image, this aberration is often disturbed and fails to be consistent across the image. As the authors mentioned, this approach is effective only when the manipulated region is relatively small allowing for a reliable global estimate. Copy-paste forgery in JPEG images has been detected by extracting the DCT block artifact grid and by identifying mismatch among the grid by Weihai Li et al. [10]. Farid et al. [15] have noticed that the color images taken from a digital camera have specific kind of correlations among the pixels, due to interpolation in the color filter array (CFA). These correlations are likely to be destroyed, when an image is tampered. They have showed that the method can reasonably distinguish between CFA and non-CFA interpolated portions of images even when the images are subjected to JPEG compression, additive noise or luminance non-linearities. But they have not discussed splice detection, when portions of different images with same CFA interpolation technique are spliced together as a composite image. Lukas et al. [16] have presented an automatic approach for the detection of tampered regions based on pattern noise, a unique stochastic characteristic of imaging sensors. The regions that lack the pattern noise are highly suspected to be forgeries. The method works in the presence of either the camera that took the image or when sufficiently many images taken by that camera are available. However this is always not possible. A semi-automatic method for the detection of image splicing based on geometry invariants and camera characteristic consistency have been proposed by Hsu [17]. The method detects Camera Response Function (CRF) for each region in an image based on geometry invariants and subsequently checks whether the CRFs are consistent with each other using cross-fitting techniques. CRF inconsistency implies splicing. The authors have used only uncompressed RAW or BMP image formats which are not always provided with all consumer level compact digital cameras, whereas our proposed approach is not restricted to the type of image format. Johnson et al. [18] have described a technique specifically designed to detect composites of images of people. This approach estimates a camera's principal point from the image of a person's eyes. Inconsistencies in the principal point are then used as evidence of tampering. As authors mentioned, the major sensitivity with this technique is in extracting the elliptical boundary of the eye. This process will be particularly difficult for low-resolution images.

Though the method proposed in this paper uses camera characteristic property; lens radial distortion for splicing detection, the method can successfully detect both copy-move and copy-paste forgery, even if they are created by using images of the same camera.

## 3. Lens Radial Distortion

Virtually all optical imaging systems introduce a variety of aberrations into an image due to its imperfections and artifacts. Lens distortion is one such aberration introduced due to geometry of camera lenses.

3.1 Background of lens radial distortion

Unlike extrinsic factors, intrinsic factors are due to camera characteristics and are specific constants to a camera, e.g. focal length, imaging plane position and orientation, lens distortion, aspect ratio etc. These are independent of position and nature of the objects captured. Lens distortion is deviation from rectilinear projection; a projection in which straight lines in a scene remain straight in an image. However, in reality almost all lenses suffer from small or large amounts of distortion. Lens radial distortion is the dominating source of mapping errors especially in inexpensive wide-angle lenses because wide-angle lens is shaped to allow a larger field of view.

Due to the shape of the lens magnification and focus is not isotropic resulting in unwanted distortions. Lens radial distortion deforms the whole image by rendering straight lines in the object space as curved lines on the film or camera sensor. Radial distortion is a non-linear transformation of the image increasing from the centre of distortion to the periphery of the image. The centre of lens distortion is a point somewhere close to the centre of the image, around which distortion due to the shape of the camera lens is approximately symmetrical. Fig. 1 shows an image of a grid and 'r' is the radius of grid. Two of the most common distortions are barrel (k>0) and pincushion (k<0) distortions (shown in fig. 2a & 2b), where k is the distortion parameter which indicates the amount of lens radial distortion (refer Section 3.2). 'r`' in fig. 2 is the deformed radii of the grid due to distortions.

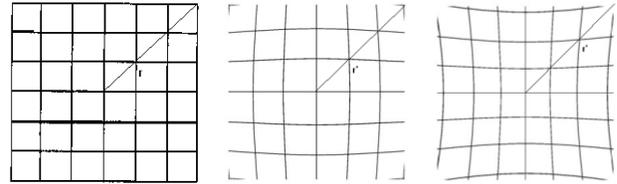

Fig. 1 No distortion　　(a) Barrel distortion　　(b) Pincushion distortion
Fig. 2  Distorted images

It is evident from fig. 2a & 2b that the amount of distortion increases with distance from centre of image to the periphery of image. We have found that this correlation among different portions of the image is disturbed in spliced images. In order to prove the integrity of an image, in this paper, we look for the consistency among the lens radial distortion parameters which are estimated from different regions of an image. This source of information is very strong, provided 3D lines are projected on the image. Thus the proposed technique works on images of city scenes, interior scenes, aerial views containing buildings and man-made structures. Apart from the lens design, the degree of radial distortion is related to focal length [19]. Usually, lenses with short focal length have a larger degree of barrel distortion, whereas lenses with long focal length suffer more from pincushion distortion. As a result, lenses from different camera leave unique imprints on the pictures being captured.

3.2  Measuring radial distortion of lenses

The lens radial distortion model can be written as an infinite series, as given below:

$$r_u = r_d\,(1 + k_1 r_d^2 + k_2 r_d^4 + \cdots) \quad \text{------------(1)}$$

with $\quad r_u = \sqrt{x_u^2 + y_u^2}\quad$ and $\quad r_d = \sqrt{x_d^2 + y_d^2}$

$r_u$ and $r_d$ are undistorted and distorted radii respectively. Radius is the radial distance $\sqrt{x^2 + y^2}$ of a point $(x, y)$ from the centre of distortion. From (1) it follows that:

$$x_u = x_d\,(1 + k_1 r_d^2 + k_2 r_d^4 + \cdots) \quad \text{--------------(2)}$$

$$y_u = y_d (1 + k_1 r_d^2 + k_2 r_d^4 + \cdots\cdots) \text{--------------(3)}$$

Centre of distortion is taken as centre of image [19]. The first-order radial symmetric distortion parameter $k_1$ is sufficient for reasonable accuracy [20]. Thus the polynomial distortion model in (2) and (3) may be simplified as

$$x_u = x_d (1 + k_1 r_d^2) \text{------------------ (4)}$$

$$y_u = y_d (1 + k_1 r_d^2) \text{------------------ (5)}$$

The parameter $k_1$ has dominant influence on the kind of radial lens distortion. If $k_1 > 0$, distortion is barrel and if $k_1 < 0$, distortion is pincushion.

### 3.3 Proposed approach

Portions of an image are correlated with each other with respect to the imaging device. Such correlations will be disturbed in spliced images. An intrinsic camera parameter viz., lens radial distortion is used for the detection of image splicing. Inconsistency in the degree of lens radial distortion across the image is the main evidence for splicing operation. In this section, a novel passive technique is described for detecting copy-paste forgery by quantitatively measuring lens radial distortion from different portions of the image using line-based calibration. Line-based calibration of lens radial distortion can be divided into three steps:

- Detection of edges with sub-pixel accuracy
- Extraction of distorted line segments
- Estimation of $k_1$ for each distorted line in an image

*Detection of edges with sub-pixel accuracy:* The first step of the calibration consists of detecting edges from an image. Since image distortion is sometimes less than a pixel, there is a need for an edge detection method with sub-pixel accuracy. We used the edge detection method proposed in [21], which is a sub-pixel refinement of the classic non-maxima suppression of the gradient norm in the direction of the gradient.

*Extraction of distorted line segments:* This calibration method relies on the constraint that straight lines in 3D must always project as straight lines in the 2D image plane, if the radial lens distortion is compensated. In order to calibrate distortion, we must find edges in the image which are most probably projections of 3D segments. Because of the distortion, a long segment may be broken into smaller segments. By defining a very small tolerance region, we can extract such distorted line segments as shown in fig. 3.

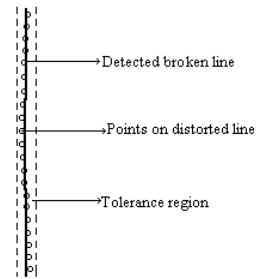

Figure. 3 Detection of broken line segments within the tolerance region

Perturbations along the lines may be generated due to low resolution of the image. Such distracted or perturbed line segments must be rejected even within the tolerance region.

*Estimation of $k_1$ for each distorted line in an image:* Measure the degree of distortion in terms of distortion parameter $k_1$ for each distorted line in the image. In order to measure the absolute deviation of a distorted line from its undistorted line, the points on a distorted line segment are used to fit a straight line using linear regression [22].

From eq. (4) and eq. (5), all '$n$' distorted points $p_{d,i}=(x_{d,i}\ y_{d,i})$ of the selected curved line are mapped to the undistorted points $p_{u,i}=(x_{u,i}\ y_{u,i})$ where ($1 \leq i \leq n$) as follows:

$$x_{u,i} = x_{d,i}(1 + k_1 r_{d,i}^2) \text{------------ (6)}$$

$$y_{u,i} = y_{d,i}(1 + k_1 r_{d,i}^2) \text{------------ (7)}$$

All '$n$' undistorted points $p_{u,i}$ should now lie on a straight line. Thus, an associated straight line '$L_n$' is represented in Hesse's normal form, it has three unknowns $n_x, n_y$ and $d_0$:

$$L_n : \begin{pmatrix} n_x \\ n_y \end{pmatrix}^T \begin{pmatrix} x \\ y \end{pmatrix} - d_0 = 0 \text{-------------- (8)}$$

To determine these unknowns with linear regression, the following expressions are calculated.

$$\overline{X} = \frac{1}{n}\sum_{i=1}^{n} x_{d,i} \qquad \overline{Y} = \frac{1}{n}\sum_{i=1}^{n} y_{d,i}$$

$$\overline{X^2} = \frac{1}{n}\sum_{i=1}^{n} (x_{d,i})^2 \qquad \overline{Y^2} = \frac{1}{n}\sum_{i=1}^{n} (y_{d,i})^2$$

$$\overline{XY} = \frac{1}{n}\sum_{i=1}^{n} x_{d,i} y_{d,i}$$

Two cases have to be distinguished:

Case 1: Let $\overline{X^2} - \overline{(X)}^2 \geq \overline{Y^2} - \overline{(Y)}^2$, then the associated straight line $L_n$ is parameterized as

$$L_n: \quad y = ax + b \quad \text{----------------- (9)}$$

with

$$a = \frac{\overline{XY} - \overline{X}\,\overline{Y}}{\overline{X^2} - \overline{(X)}^2} \qquad b = \frac{\overline{X^2}\,\overline{Y} - \overline{X}\,\overline{XY}}{\overline{X^2} - \overline{(X)}^2}$$

and the three unknowns are as follows:

$$n_x = \frac{-a}{\sqrt{a^2 + 1}} \qquad n_y = \frac{1}{\sqrt{a^2 + 1}} \qquad d_0 = \frac{b}{\sqrt{a^2 + 1}}$$

Case 2: Let $\overline{X^2} - \overline{(X)}^2 < \overline{Y^2} - \overline{(Y)}^2$, then the parameterization of the associated straight line $L_n$ changes to:

$$L_n: \quad x = cy + d \quad \text{----------------- (10)}$$

with

$$c = \frac{\overline{XY} - \overline{X}\,\overline{Y}}{\overline{Y^2} - \overline{(Y)}^2} \qquad d = \frac{\overline{X^2}\,\overline{Y} - \overline{X}\,\overline{XY}}{\overline{Y^2} - \overline{(Y)}^2}$$

In this case the three unknowns of eq. (8) are

$$n_x = \frac{1}{\sqrt{c^2 + 1}} \qquad n_y = \frac{-c}{\sqrt{c^2 + 1}} \qquad d_0 = \frac{d}{\sqrt{c^2 + 1}}$$

Now an associated straight line $L_n$ is found, which is a function of $k_1$ and the points $p_{d,i}$.

Thus a cost function with the residual errors $\varepsilon_n$ of eq. (8) is formulated as:

$$\varepsilon_i = \begin{pmatrix} n_x \\ n_y \end{pmatrix}^T \begin{pmatrix} x_{u,i} \\ y_{u,i} \end{pmatrix} - d_0$$

Substitute $x_{u,n}$ and $y_{u,n}$ from eq. (6) and (7)

$$\varepsilon_i = \begin{pmatrix} n_x \\ n_y \end{pmatrix}^T \begin{pmatrix} x_{d,i}(1 + k_1 r_{d,i}^2) \\ y_{d,i}(1 + k_1 r_{d,i}^2) \end{pmatrix} - d_0$$

Where $k_1$ is selected to minimize $\sum_{i=1}^{n} \varepsilon_i^2$.

This cost function is a non-linear function of $x_{d,i}$ and $y_{d,i}$ of a curved line or distorted line. The deviation of points $(x_{d,i}\ y_{d,i})$ from their original positions $(x_{u,i}\ y_{u,i})$ is used to estimate the amount of distortion. The distortion error is the sum of squares of distances from the points to the straight line. To estimate distortion parameter $k_1$ the sum of squares is minimized using the iterative Levenberg-Marquardt method through *lsqnonlin* function found in MATLAB. Thus we obtain unique $k_1$ for each distorted line segment in an image depending on the amount of distortion.

## 4. Experimental Setup

Three sets of experiments were performed. The first set of experiments aims at technical investigation and calibration of lens radial distortion for different consumer level compact digital cameras. The second set of experiments conducted on synthetic images to show how to use radial distortion parameter as a feature to detect image splicing. The third set of experiments study the performance of proposed features on real images.

4.1 Analysis of lens radial distortion for different cameras

To analyze the behavior of intrinsic radial distortion parameter across the image, we have used 6 digital cameras of recent models from four different manufacturers. The configurations of the cameras are given in table 1.

Table 1. Cameras used in experiments and their properties

| Camera Brand | Focal Length(mm) | Optical Zoom | Resolution (Mega Pixel) |
|---|---|---|---|
| Sony DSC-W35 | 38-114 | 3x | 7.2 |
| Sony DSC-W220 | 29.7-118.8 | 4x | 12.1 |
| Canon A550 | 35-140 | 4x | 7.1 |
| Casio EX-Z75 | 38-114 | 3x | 7.1 |
| Nikon S550 | 36-180 | 5x | 10 |
| Sony DSC-W180 | 35-105 | 3x | 10.1 |
| Note: The focal length is equivalent to a 35mm film camera | | | |

Fig. 4 shows the checker-board with 9 by 12 square grids, generated manually without any distortions. Fig. 5 shows the extracted straight lines from fig. 4. The lens radial distortion parameter $k_1$ is computed for each straight line (ref section 3.3) and is reported zero for all straight lines and the same is shown as a graph in figure 6. All through the experiments, it is assumed that the image centre as (0,0) and the horizontal and vertical coordinates are normalized so that the maximum of the dimensions is in the range (-1,1). Most consumer digital cameras are equipped with an optical zoom lens. The lens radial distortion parameters change with focal length, which usually varies from barrel at the wide end to pincushion at the tele end. In this section we investigate the impact of optical zoom on the behaviour of radial distortion parameter across the image.

To study the actual behaviour of radial distortion parameter $k_1$ across the image, images of the same scene are acquired from different cameras. The checker board (shown in fig. 4) is captured, approximately with same distance, position and orientation of the camera. The images were taken with no flash, auto-focus, JPEG format and other default settings. The sample images captured by Sony DSC-W35 camera with different zoom levels is shown in fig. 7. Fig. 8 is the corresponding edge images of each image in fig. 7. Since the radial distortion is approximately symmetric we have drawn the graph of $k_1$ for vertical lines which lies in the right of image centre. Figures 9-14 show the behaviour of radial distortion parameter $k_1$ across the image for various cameras at different zoom levels. It is clear that no camera is ideal. All cameras have noticeable amount of radial distortion. It is also evident from the graph that the degree of radial distortion increases with the distance from centre of the image. We can also observe that the radial distortion parameter changes with zoom and most of the cameras vary from barrel at the wide end to pincushion at the tele end (refer fig. 9, 10, 12 and 14).

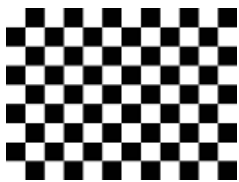
Fig. 4 Checker-board
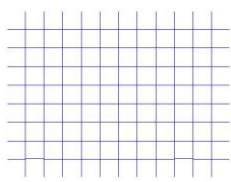
Fig. 5 Extracted edges

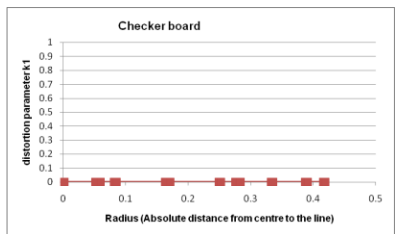
Fig. 6 Graph of $k_1$ for lines in fig. 5

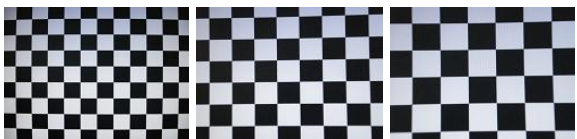
Fig. 7 Checker-board image captured by Sony DSC-W35 camera

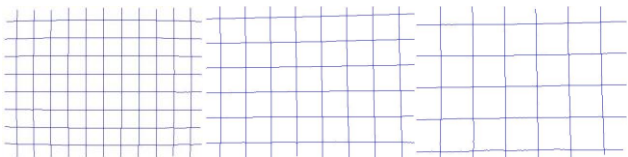
Fig. 8 Extracted edges from fig 7

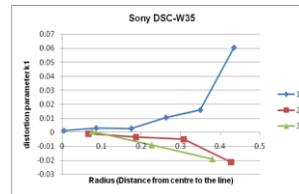
Fig. 9
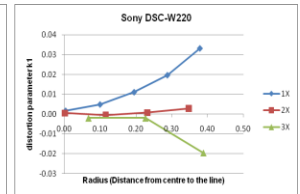
Fig. 10

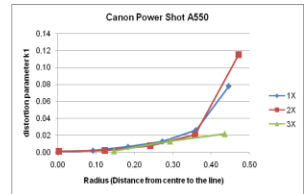
Fig. 11
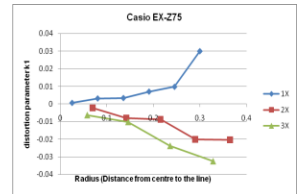
Fig. 12

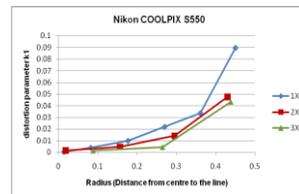
Fig. 13
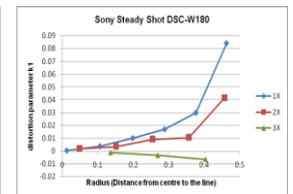
Fig. 14

Figs. 9-14 Behaviour of lens radial distortion parameter $k_1$ across the image for various cameras at different zoom levels

### 4.2 Experiments on Synthetic Images

In this section we describe, how to use lens radial distortion parameter for the detection of image splicing. If two image regions belong to the same image (untampered) then the radial distortion of extracted line segments should behave as expected (explained in Section 4.1). That is lines that are more or less equidistant will suffer from more or less equal amount of radial distortion. Also radial distortion is uniform, barrel or pincushion.
Usually the composites are created in three ways:
  (i)   Image splicing on the same image (copy-move forgery)
  (ii)  Two or more images of the same camera (may be captured at different zoom) are used to create a composite image
  (iii) Two or more images of the different cameras (irrespective of the camera make or model) are used to create a spliced image

Hence, we used three different cameras, in which two are from same manufacturer, so that all types of splicing described above can be carried out. Fig. 15 is the image acquired from Sony DSC-W35 and the extracted straight lines are shown in fig. 16. Table 2 shows the distances (column 2) of straight lines from the image centre and its corresponding values of $k_1$ (column 3). Line 1 is the left most and line 6 is the right most line. You can observe that the values of $k_1$ gradually increase from image centre to its periphery. Similar observations on $k_1$ were already noticed in Section 4.1 (refer graph of $k_1$ values of various cameras in figures 9-14). This consistency will be disturbed in case of copy-move or copy-paste forgery. Inconsistency in lens radial distortion can be detected if one of the two conditions is not met: (a) Amount of radial distortion is symmetric and increases with the distance from centre of the image and (b) Sign of $k_1$ should be same for all lines throughout the image.

In the untampered image (fig. 15), all bottles are of different colours. The composites were created by copying and pasting a bottle over another bottle. The different cases of image splicing described in (i) to (iii) are shown in figure 17-20. The original images were captured at different zoom levels and some post-processing operations like rotation and scaling have also been done while pasting a portion. The values of $k_1$ for each line from left to right for images in figure 17-20 are listed in tables 3-6 respectively. Inconsistency is highlighted in all the tables. When a bottle on right is pasted over a bottle on left (ssown in fig. 17) then the radial distortion of the corresponding edges would change to barrel (pincushion) where as the actual distortion in untampered image would have been pincushion (barrel). This can be noticed in table 3. Image in figure 18 has been spliced by replacing the middle bottle by a bottle on the right side of another image and $k_1$ values for all extracted lines is given in table 4. $k_1$ is negative for line 3 and positive for all other lines indicating inconsistency of type (b). Observe that $k_1$ of line 4 is greater than that of line 5, this implies inconsistency of type (a). Thus splicing has been successfully detected.

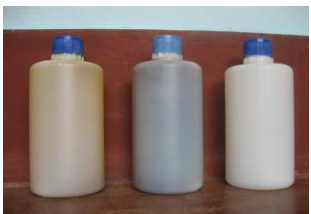
Fig. 15 An image of Sony DSC-W35

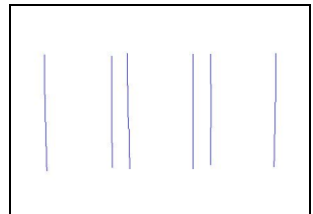
Fig. 16 Extracted Edges

Table 2. Estimated $k_1$ for each line (from left to right) of fig. 16

| Straight line no. | Distance from centre | Distortion parameter $k_1$ |
|---|---|---|
| 1 | -0.4095 | 0.01439 |
| 2 | -0.1727 | 0.00455 |
| 3 | -0.1139 | 0.00065 |
| 4 | 0.1181 | 0.00071 |
| 5 | 0.1809 | 0.00478 |
| 6 | 0.4112 | 0.01485 |

In real cases, the creation of composites is commonly accompanied with subsequent operations like JPEG compression, contrast/brightness adjustment, color changing, blurring, rotation, resizing etc. to hide the obvious traces of tampering. Since the proposed splicing detection method works on images consisting of straight edges, contrast, brightness and color manipulations will not affect edge detection, unless the object color and the background color are indistinguishable. It is evident from the above experiments that the proposed approach is robust to rotation, as distortion is same even if lines are rotated. We observed that JPEG compression, blurring and resizing operations affects on the performance of the proposed method. Experiments show that JPEG compression with a quality factor of 5 (out of 10) also can detect straight lines with reasonable accuracy sufficient to estimate $k_1$. Resizing is the most common operation performed while creating composites, in order to match with the size of the host image. The proposed approach is robust to resizing and blurring provided the extracted lines continue to be unperturbed.

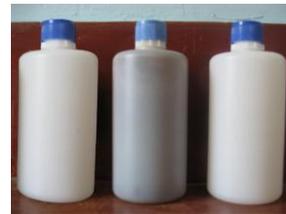
Fig. 17 Copy-move or Region-duplication

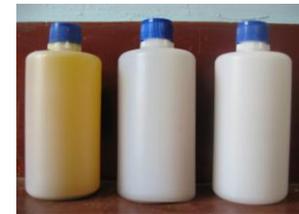
Fig. 18 Composite of 2 images captured by cameras of different make

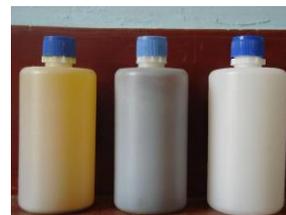
Fig. 19 Composite of two images captured by same camera

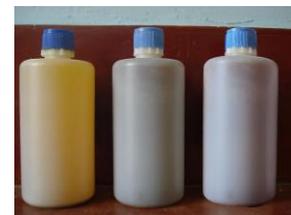
Fig. 20 Composite of two images captured by cameras of different model

Table 3. Estimated $k_1$ for each line (from left to right) in fig. 17

| Straight line no. | Distance from centre | Distortion parameter $k_1$ |
|---|---|---|
| **1** | **-0.4736** | **-0.00371** |
| **2** | **-0.1981** | **-0.04611** |
| 3 | -0.144 | 0.000923 |
| 4 | 0.1193 | 0.000872 |
| 5 | 0.1934 | 0.003192 |
| 6 | 0.4614 | 0.048074 |

Table 4. Estimated $k_1$ for each line (from left to right) in fig. 18

| Straight line no. | Distance from centre | Distortion parameter $k_1$ |
|---|---|---|
| 1 | -0.4688 | 0.038074 |
| 2 | -0.2033 | 0.004531 |
| **3** | **-0.1447** | **-0.00172** |
| **4** | **0.1285** | **0.01014** |
| 5 | 0.1934 | 0.0043 |
| 6 | 0.4614 | 0.034251 |

Table 5. Estimated $k_1$ for each line (from left to right) in fig. 19

| Straight line no. | Distance from centre | Distortion parameter $k_1$ |
|---|---|---|
| 1 | -0.4704 | -0.00077 |
| 2 | -0.199 | -0.00169 |
| 3 | -0.1337 | -0.01394 |
| 4 | 0.1247 | -0.0128 |
| **5** | **0.1963** | **0.002672** |
| **6** | **0.4754** | **0.038046** |

Table 6. Estimated $k_1$ for each line (from left to right) in fig. 20

| Straight line no. | Distance from centre | Distortion parameter $k_1$ |
|---|---|---|
| 1 | -0.4798 | 0.107796 |
| 2 | -0.2111 | 0.001981 |
| 3 | -0.1464 | 0.001776 |
| 4 | 0.1118 | 0.0008 |
| **5** | **0.1782** | **-0.00331** |
| **6** | **0.4454** | **0.019655** |

4.3 Experiments on real images

The non-availability of the suitable data set
for examining the proposed method led us to create our own Spliced Image Data Set (SIDS). However, we have also experimented with a few images from the database provided by Hsu [17]. In order to compute the splice detection rate, we have created 100 spliced images from 350 authentic images. Authentic images were taken with our 6 consumer level compact digital cameras (mentioned in Table 1) and 50 images are downloaded from internet. 50 images were captured from each of 6 cameras in JPEG format with dimensions ranging from 1632x1224 to 3072x2304. These images mainly contain indoor scenes like computers, boards, tables, library, photo frames etc. Some images contain outdoor scenes like buildings, shopping complex etc. We created spliced images from the authentic image set using Adobe Photoshop. In order to hide the traces of tampering, some post processing operations like resizing and rotation were performed.

As a first step, all distorted line segments from each spliced image are detected as described in Section 3.3. Further, for accurate comparison of distortion parameter $k_1$ for each detected line in an image, the length of those lines must be equal. All lines detected in an image are trimmed to that of the shortest line (at least $1/3^{rd}$ of the image height). The detection rate for our spliced image dataset is found as 86%. Some sample spliced images of our dataset and their extracted line segments are shown in fig. 21. The pasted portion is indicated by red lines.

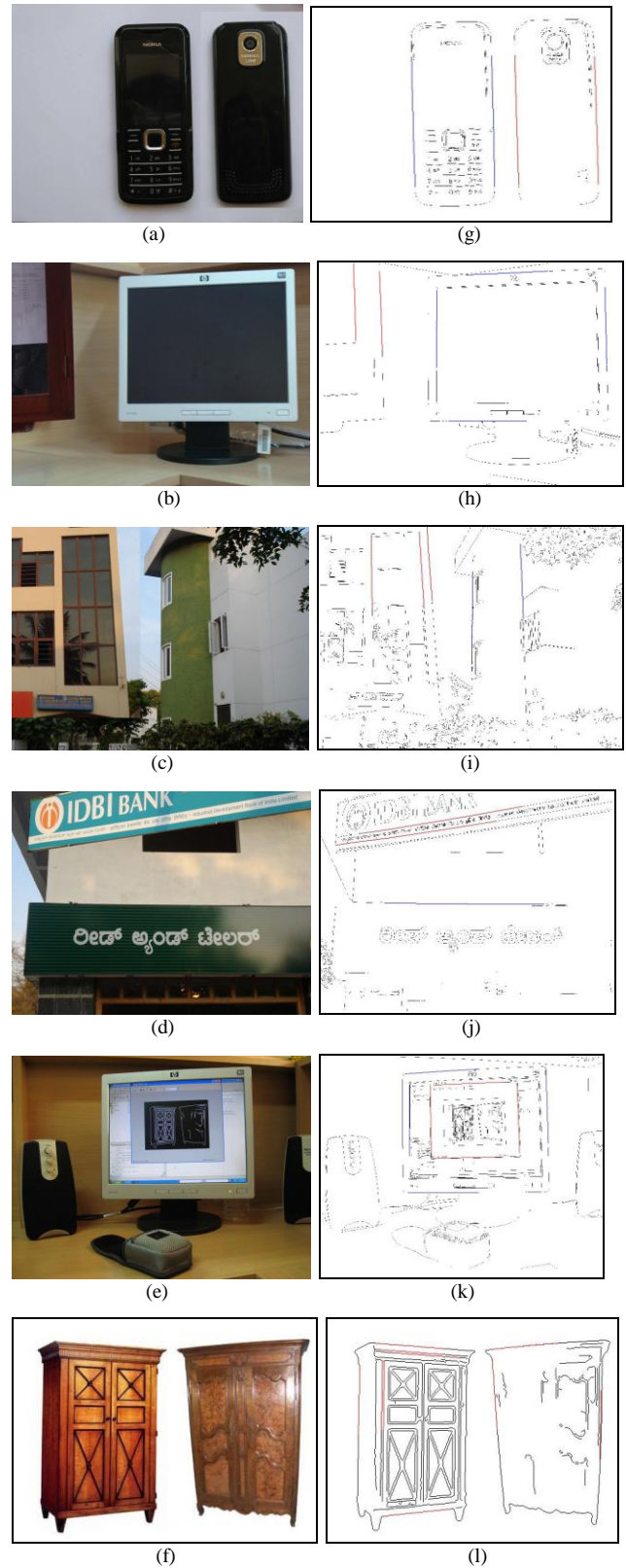

Fig. 21. (a)-(f) Sample spliced images and (g)-(l) are their extracted edges respectively

We have also verified our proposed approach with images selected from Columbia Image Splicing Detection Evaluation dataset. However, we have selected those spliced images that contain straight edges. Some sample images are shown in fig. 22. The pasted portion is indicated by red lines.

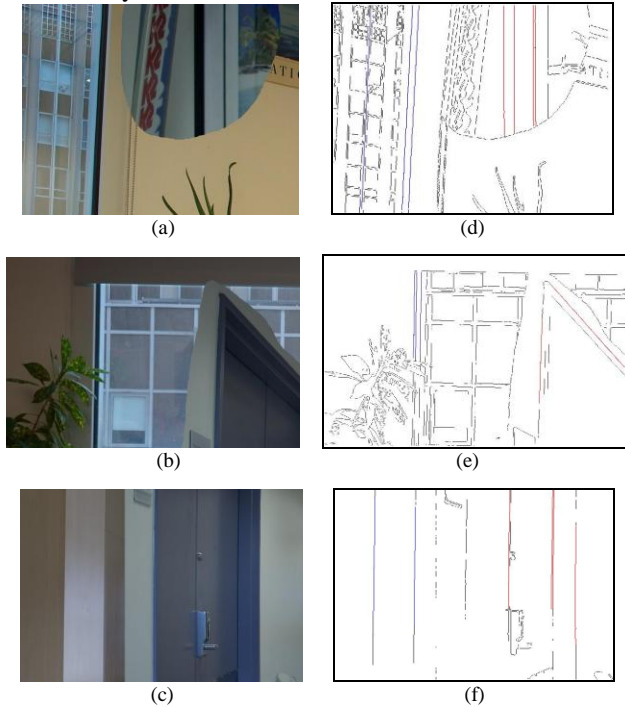

Figs. 22. (a)-(c) Sample images from database [17] and (d)-(f) are their extracted edges respectively

## 5. Discussion and future work

The proposed approach detects splicing of images when straight edges are available, but the main difficulty lies in the extraction of such straight edges. Sometimes the low quality image will generate perturbations along the straight lines which results in wrong estimation of radial distortion parameter $k_1$. Because of low resolution, an authentic image in fig. 23 is detected as spliced image (perturbations may be visible in enlarged version).

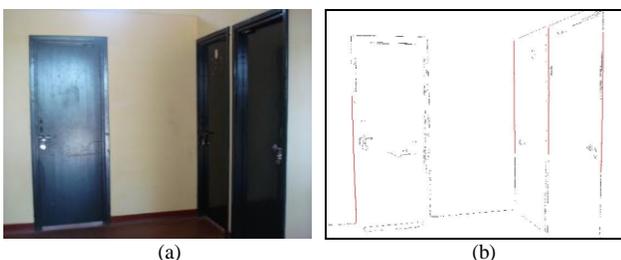

Fig. 23. (a) A low quality image and (b) shows the extracted perturbed lines

We point out that our method may not provide sufficiently conclusive statistical evidence regarding spliced portions of the image. It means the proposed approach may fail to detect a spliced portion if two images have same kind of distortion are spliced together and the position of copied and pasted portion is same with respect to the centre of image. Also, more research and analysis are needed to determine the influence of centre of distortion and the length of lines.

Though the proposed method works for the images from lower-end consumer level digital cameras taken with default settings and most commonly available JPEG image format, we need at least two or more straight lines with the minimum length of $1/3^{rd}$ of the image height from different regions in order to prove the integrity of the image. To summarize, the proposed method works best on images with lot of lines and of significant lengths, such as images of city scenes, interior scenes, aerial views containing buildings and man-made structures. Probably the experimental results can be improved by using a more sophisticated method to estimate the lens radial distortion across the image. We expect the proposed technique, when integrated with other available splicing detection methods [14-18], to become more effective in exploring digital forgeries.

It is interesting to address malicious attacks intended to interfere the detection algorithm, such as adding or removing radial distortion from an image [23, 24]. Since it is possible to manipulate the distortion parameters of an image globally, these alterations will not affect the performance of our method if done on spliced image.

## 6. Conclusion

Portions of the image are correlated with each other with respect to the imaging device. Such correlations will be disturbed in spliced images. We have used an intrinsic camera parameter, namely lens radial distortion, for the detection of image splicing. Inconsistency in the degree of lens radial distortion across the image is the main evidence for the detection of spliced images. In this paper we propose a novel passive technique (with no watermark or signature) for detecting copy-paste forgery by quantitatively measuring lens radial distortion from different portions of the image using line-based calibration.

Experiment in section 4.1 shows that most consumer level digital cameras have small or large amount of lens radial distortion at different zoom levels. Experimental set up in section 4.2 demonstrates how efficiently the lens radial distortion parameter $k_1$ may be used for the detection of image splicing and the experimental results in section 4.3 shows that our method works well in case of real images. The primary contribution of our work is that we examine the use of inherent lens distortion as a unique imprint on the images for the detection of image splicing.

**H.R. Chennamma** received her graduate degree in Computer Applications with distinction in the year 2003, Vishwesharaiah Technological University, India. She was a Project Trainee for a year at the National Aerospace Laboratory (NAL), Bangalore, India. She served as a software engineer for a year in a multinational software company, Bangalore, India. Mrs. Chennamma is now a Senior Research Fellow (SRF) in National Computer Forensic Laboratory, Ministry of Home Affairs, Government of India, Hyderabad since 2005. Subsequently, she was awarded Ph.D. program fellowship at the Department of Computer Science, University of Mysore in the year 2006. Mrs. Chennamma is the recipient of "Best Scientific Paper Award" in the All India Forensic Science Conference, Kolkata, India in the year 2007. Her current research interests are Image Forensics, Pattern Recognition and Image Retrieval.

**Dr. Lalitha Rangarajan** has two graduate degrees to her credit, one in Mathematics from University of Madras, India (1980) and the other in Industrial Engineering (Specialization: Operational Research) from Pardue University, USA during (1988). She has taught Mathematics briefly at graduate level, for 5 years during 1980 to 1985 in India. She joined Department of Computer Science, University of Mysore, to teach graduate students of the Department, in 1988, where she is currently working as a Reader. She completed Ph.D. in Computer Science in the area of pattern recognition in 2004. She is presently working in the areas of Feature Reduction, Image Retrieval and Bioinformatics.